\documentclass{article}
\usepackage{PRIMEarxiv}
\usepackage[utf8]{inputenc} 
\usepackage[T1]{fontenc}    
\usepackage[utf8]{vietnam}
\usepackage[english]{babel}
\usepackage[safe]{tipa}
\usepackage{ragged2e}

\usepackage{hyperref}       
\usepackage{url}            
\usepackage{booktabs}       
\usepackage{amsfonts}       
\usepackage{amsmath}
\DeclareMathOperator{\diag}{diag}
\DeclareMathOperator{\sat}{sat}
\newcommand{\norm}[1]{\left\lVert#1\right\rVert}
\usepackage{nicefrac}       
\usepackage{microtype}      
\usepackage{lipsum}
\usepackage{graphicx}
\graphicspath{{figures/}}     
\usepackage{caption}
\usepackage{subcaption}
\usepackage{multirow}

\usepackage[final]{pdfpages}
\usepackage{tabularx}
\usepackage{tkz-tab}
\usepackage{framed,multicol}
\usepackage[framemethod=tikz]{mdframed}
\usepackage{xcolor}

\usepackage{wasysym}
\usepackage{pgfplots}
\usepackage{tikz}
\usepackage{pgfplotstable}
\usepackage{filecontents}

\usepgfplotslibrary{colorbrewer}
\pgfplotsset{compat=1.18, cycle list/Set1-8} 
\usetikzlibrary{pgfplots.statistics, pgfplots.colorbrewer} 

\usetikzlibrary{positioning, fit, calc, arrows.meta}  
\tikzset{block/.style={draw, thick, text width=2cm ,minimum height=1.3cm, align=center},   
	line/.style={-latex}     
} 

\tikzset{blocktext/.style={draw, thick, text width=5.2cm ,minimum height=1.3cm, align=center},   
	line/.style={-latex}     
}

\tikzset{font=\footnotesize}

\title{How Tuning, Reference Preview, Actuator Modeling, and Fallback Policies Change Quadrotor Controller Comparisons: A Reproducible Simulation Study}

\author{
	Xuan-Quy Dao\hspace{1mm}\href{https://orcid.org/0000-0001-5415-7538}
	{\includegraphics[scale=0.06]{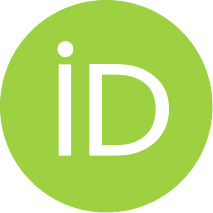}} \\
	Eastern International University\\
	Binh Duong, Vietnam \\
	\texttt{quy.dao@eiu.edu.vn} \\
	\And
Le-Minh Nguyen\hspace{1mm}\href{https://orcid.org/0009-0002-0384-3938}
{\includegraphics[scale=0.06]{orcid.pdf} } \\
	Eastern International University\\
	Binh Duong, Vietnam \\
	\texttt{minh.nguyenle@eiu.edu.vn} \\
	\And
	Ngoc-Bich Le\hspace{1mm}\href{https://orcid.org/0000-0001-7431-0157}
	{\includegraphics[scale=0.06]{orcid.pdf}}
	\thanks{Corresponding author: Ngoc-Bich Le (\texttt{lengocbich@iuh.edu.vn}).} \\
	Industrial University of Ho Chi Minh City\\
	Ho Chi Minh City, Vietnam \\
	\texttt{lengocbich@iuh.edu.vn}
}

\hypersetup{
pdftitle={How Tuning, Reference Preview, Actuator Modeling, and Fallback Policies Change Quadrotor Controller Comparisons: A Reproducible Simulation Study},
hidelinks,
pdfsubject={Quadrotor controller benchmarking},
pdfauthor={Xuan-Quy Dao, Le-Minh Nguyen, Ngoc-Bich Le},
pdfkeywords={quadrotor, trajectory tracking, model predictive control, controller benchmarking, actuator dynamics, fallback control, computation delay, reproducibility},
}

\begin{document}
\maketitle

\begin{abstract}

Quadrotor controller comparisons commonly report a single tracking error, although the ranking also depends on tuning effort, reference information, actuator prediction, optimizer fallback, and command timing. We quantify these effects for cascade PID, hover LQR, boundary-layer sliding-mode control (BL-SMC), linear MPC with LQR fallback (LMPC-F), a preview-enabled LMPC-F, and a geometric controller on a common nonlinear plant with rotor gyroscopic coupling, motor lag, and saturation. A primary study (556 episodes at 50~Hz) combines matched twelve-candidate tuning, 400 paired evaluation episodes, and a 48-cell MPC factorial; a confirmation study (1462 episodes) repeats the evaluation on 50 fresh realizations with tuning budgets of 12, 24, and 48 candidates. Matched tuning changed the mean-RMSE ordering of the fixed baseline settings, mainly by moving BL-SMC from last to first. The ordering recurred on fresh realizations and did not change with 48 candidates, although BL-SMC's lead there depends on a wider search range for one of its gains. Reference preview reduced MPC tracking error and removed fallback activity on the figure-eight task. In single abrupt-step runs, completion depended jointly on fallback policy, predictor, and actuator authority. Replaying measured command-call delays terminated three of four nominal MPC configurations on a desktop and all four on a Raspberry Pi~5, whereas capping the solver at 100 iterations removed every deadline miss in these profiles. We propose a reporting checklist that adds tuning effort, failure rates, fallback activity, applied-input variation, and command timing to tracking accuracy.

\end{abstract}


\section{Introduction}\label{sec:intro}

Quadrotor trajectory tracking is inherently coupled with attitude regulation because translational motion is generated by reorienting the thrust vector, while achievable performance is constrained by motor dynamics, thrust saturation, and power limitations~\cite{mahony2012,faessler2017}. Controller comparisons are therefore a recurring theme in the literature, from early proportional--integral--derivative (PID), linear--quadratic (LQ), backstepping, and sliding-mode studies~\cite{bouabdallah2004,bouabdallah2005,bouabdallah2007} to recent comparisons of linear and nonlinear model predictive control (MPC) and differential-flatness-based control~\cite{kamel2017,sun2022}. Surveys of multirotor, PID, and MPC control~\cite{nascimento2019,lopez2023,nguyen2021mpc} and of agile autonomous flight~\cite{hanover2024} summarize the broad design space.

The observed outcome of such a comparison, however, depends on more than the feedback law. It is also influenced by (i) the effort devoted to tuning each method, (ii) the availability of future reference information, (iii) whether motor dynamics are represented in the predictor, (iv) the command applied when the optimizer fails or returns an unacceptable solution, and (v) whether each command is available within the sampling period. The last two factors are especially important for optimization-driven controllers, since the system may receive a mix of accepted optimizer results and fallback actions, and each optimizer call has a non-negligible computation latency~\cite{findeisen2004,wang2010}. For these reasons, simply reporting an average tracking error can hide these influences.

This paper examines these effects in a controlled simulation study of cascade PID control, a hover linear--quadratic regulator (LQR), boundary-layer sliding-mode control (BL-SMC), and linear MPC with LQR fallback (LMPC-F). A preview-enabled LMPC-F and a geometric $SO(3)$ controller are included in a confirmation study using fresh realizations. All controllers are evaluated on the same nonlinear plant, with a common rotor allocator and a 50~Hz update rate, while retaining their method-specific reference construction, feedforward terms, and inner-loop structure. The study addresses three questions:
\begin{enumerate}
	\item How does a matched candidate-count tuning protocol affect the accuracy ordering, and does the resulting ordering persist on fresh realizations and under larger candidate-count budgets?
	\item How do reference preview, motor-state prediction, fallback projection, and tilt-constraint placement individually and jointly affect LMPC-F under smooth and discontinuous references?
	\item How sensitive are the closed-loop results to measured command-call delays and missed deadlines?
\end{enumerate}

The main contributions of this work are as follows.
\begin{itemize}
	\item A paired fixed-versus-tuned evaluation over 50 randomized realizations demonstrates that a matched candidate-count tuning protocol changes the ordering of the four controllers by mean root-mean-square error (RMSE), and that the accuracy gains are accompanied by controller-dependent changes in actuation smoothness.
	\item A factorial analysis of LMPC-F design choices shows that reference preview has a stronger effect than predictor choice during a smooth maneuver. In single deterministic runs of an abrupt step, however, projected fallback completed only in combination with preview, motor prediction, and reduced actuator authority; this pattern describes the tested trajectories and has not been replicated over uncertainty realizations.
	\item An analysis of LMPC-F applied-input variation shows that most of the large variation occurs within fallback intervals rather than at transitions between QP and fallback operation, and identifies a possible cause in how the previous request is carried over after a fallback.
	\item A confirmation study on 50 fresh realizations extends the evaluation to a geometric $SO(3)$ controller, preview-enabled MPC, and nested tuning budgets of 12, 24, and 48 candidates. It reproduces the change in the mean-RMSE ordering, shows that the ordering is maintained when the candidate-count budget is quadrupled, and finds that reference preview removes the large variation in the MPC input. A sensitivity check shows that BL-SMC's first place on the fresh set depends on the wider search range used for its vertical switching gain.
	\item A replay using recorded command-call timing traces on a desktop computer and a Raspberry Pi~5 demonstrates that numerical acceptance of a control solution does not guarantee its timely delivery to the actuators, and that an iteration cap tied to the deadline removed the resulting failures in the tested nominal timing profiles. These results provide the basis for a reporting checklist in Table~\ref{tab:checklist} for future quadrotor controller comparisons.
\end{itemize}

All configurations, raw logs, and timing traces are publicly released to support reproducibility (see the Data and code statement). An independent rerun with the same pinned dependencies reproduces every non-timing array to within $10^{-12}$, and the confirmation study reproduces the non-timing metrics of all 96 shared development episodes exactly. Across processor architectures, trajectories close to the stability boundary can differ (Section~\ref{sec:timing-results}).

The remainder of the paper is organized as follows. Section~\ref{sec:related} reviews related controller benchmarks. Sections~\ref{sec:model} and~\ref{sec:ctrl} describe the quadrotor model and controller implementations, respectively. Section~\ref{sec:design} presents the experimental protocol, and Section~\ref{sec:results} reports the results. Section~\ref{sec:discussion} discusses verification, practical implications, and limitations. Finally, Section~\ref{sec:conclusion} concludes the paper.

\section{Related Work}
\label{sec:related}

\subsection{Quadrotor Controller Comparisons}

Bouabdallah et al. compared PID and LQ control on an indoor micro quadrotor, where the observed differences depended on the model used for controller design~\cite{bouabdallah2004}; they later compared backstepping and sliding-mode control~\cite{bouabdallah2005} and extended backstepping to full position control~\cite{bouabdallah2007}. Geometric tracking on $SE(3)$~\cite{lee2010} and differential-flatness-based control~\cite{mellinger2011,faessler2018} provide high-performance baselines beyond hover linearization. Kamel et al.~\cite{kamel2017} compared linear and nonlinear MPC for trajectory tracking, and Sun et al.~\cite{sun2022} found that the relative merit of nonlinear MPC and flatness-based control depends on reference feasibility and on how the inner loop and rotor drag are handled. Time-optimal contouring MPC~\cite{romero2022,krinner2024} and data-driven or learned aerodynamic models~\cite{torrente2021} extend the operating envelope further. These studies motivate the present focus on implementation choices rather than on the feedback law alone.

\subsection{Benchmark Frameworks}
Martini et al.~\cite{martini2023} evaluated seven controller architectures over six missions on a nonlinear quadrotor model with motor lag and saturation. Safe-Control-Gym~\cite{yuan2022} provides symbolic dynamics, constraints, and repeatable disturbances for cart-pole and reduced-dimensional quadrotor tasks, and gym-pybullet-drones~\cite{panerati2021} offers a physics-based multi-agent environment for learning-based control. Zhang et al.~\cite{zhang2025} introduced a RotorPy-based suite for adaptive control under disturbances, uncertainty, and delayed actuation. Table~\ref{tab:framework} summarizes their documented features. Because plant models, interfaces, and objectives differ, the numerical results are not directly comparable with those reported here.

\subsection{MPC Implementation Issues}
Stability and feasibility of constrained MPC are classically analyzed under exact, timely solutions~\cite{mayne2000,rawlings2017}. In practice, real-time MPC relies on fast quadratic-programming (QP) solvers such as OSQP~\cite{stellato2020}, qpOASES~\cite{ferreau2014}, or acados~\cite{verschueren2022}, with bounded iterations and early termination~\cite{wang2010} or reduced-order prediction models~\cite{bangura2014}. Soft state constraints preserve feasibility, and with an exact penalty the original hard constraint is still satisfied whenever it is feasible~\cite{kerrigan2000,zeilinger2014}; and computation delay degrades closed-loop behavior unless it is explicitly accounted for~\cite{findeisen2004}. How fallback policies, solver acceptance rules, and measured delays jointly shape quadrotor tracking is rarely reported; this study makes these elements explicit and measurable. Related work in machine learning has shown that published comparisons can reverse once every method receives the same tuning budget~\cite{melis2018,lucic2018} and that reported results are sensitive to implementation and evaluation choices~\cite{henderson2018}, while random and stratified sampling provide strong baselines for finite-budget hyperparameter optimization~\cite{bergstra2012,mckay1979}. The present protocol applies related concerns to classical and optimization-based flight-control benchmarking.

\begin{table*}[!t]\caption{Documented features of related quadrotor benchmark frameworks and of the present study.}\label{tab:framework}\centering\footnotesize
	\begin{tabular}{p{2.4cm}p{3.4cm}p{4.2cm}p{4.6cm}}\toprule
		Framework & Plant & Focus & Evaluation design\\\midrule
		Martini et al.~\cite{martini2023} & Nonlinear quadrotor, motor lag and saturation & Linear and nonlinear controllers & Seven architectures, six missions\\
		Safe-Control-Gym~\cite{yuan2022} & Cart-pole, 1D/2D quadrotor & Constraints, disturbances, safe learning & Reduced-dimensional tasks\\
		gym-pybullet-drones~\cite{panerati2021} & PyBullet multi-rotor & Learning-based, multi-agent control & Gym environments\\
		Zhang et al.~\cite{zhang2025} & RotorPy quadrotor & Adaptive vs. non-adaptive control, stress, delays & Control-library comparison\\
		This study & 16-state plant with rotor lag, saturation, gyroscopic coupling & Tuning budget, preview, predictor, fallback, tilt constraints, timing & Paired randomized evaluation, factorial MPC contrasts, measured-delay replay\\
		\bottomrule\end{tabular}\end{table*}

\section{Quadrotor and Actuator Model}\label{sec:model}
\subsection{Frames and Rigid-Body Dynamics}
The plant is a ``+''-configuration quadrotor with physical coefficients from~\cite{luukkonen2011}, first-order rotor lag, and finite rotor-speed ceilings (Table~\ref{tab:parameters}). The inertial and body frames are $\mathcal F_E$ and $\mathcal F_B$. The inertial position is $\boldsymbol\xi=[x,y,z]^\top$ with $z$ pointing upward, the Euler angles are $\boldsymbol\eta=[\phi,\theta,\psi]^\top$, the body rates are $\boldsymbol\nu$, and the rotor speeds are $\omega_i$. The rigid state is
\begin{equation}
	\mathbf X=[\boldsymbol\xi^\top,\dot{\boldsymbol\xi}^\top,\boldsymbol\eta^\top,\boldsymbol\nu^\top]^\top\in\mathbb R^{12}.
\end{equation}
Let $\mathbf R=R_z(\psi)R_y(\theta)R_x(\phi)$ and $\mathbf I=\diag(I_{xx},I_{yy},I_{zz})$, let $A=0.25\mathbf I_3$~kg/s denote the linear drag matrix, and let $\Omega_r=\omega_1-\omega_2+\omega_3-\omega_4$ denote the signed rotor-speed sum. The rotation matrix $\mathbf R$ maps vectors from the body frame to the inertial frame. Since the thrust force acts along the positive body $z$-axis, its inertial-frame representation is $T\mathbf R\mathbf e_3$. The translational and rotational dynamics are therefore written as
\begin{align}
	m\ddot{\boldsymbol\xi}&=-mg\mathbf e_3+T\mathbf R\mathbf e_3-A\dot{\boldsymbol\xi}+\mathbf F_d,\label{eq:trans}\\
	\mathbf I\dot{\boldsymbol\nu}&=-\boldsymbol\nu\times\mathbf I\boldsymbol\nu-I_M(\boldsymbol\nu\times\mathbf e_3)\Omega_r+\boldsymbol\tau,\\
	\dot{\boldsymbol\eta}&=\mathbf W^{-1}(\boldsymbol\eta)\boldsymbol\nu,
\end{align}
where $T$ denotes the total thrust magnitude, $\mathbf F_d$ represents external force disturbances (no external torque disturbance is applied), $\mathbf e_3=[0,0,1]^\top$, and
\begin{equation}\mathbf W^{-1}(\boldsymbol\eta)=\begin{bmatrix}1&\sin\phi\tan\theta&\cos\phi\tan\theta\\0&\cos\phi&-\sin\phi\\0&\sin\phi/\cos\theta&\cos\phi/\cos\theta\end{bmatrix}.\end{equation}
The signed rotor-speed combination $\Omega_r$ follows the rotor numbering and spin directions stated in Section~\ref{sec:rotor}. The gyroscopic term is implemented with this convention in both the nonlinear plant and the verification tests, so its sign should be read together with that rotor convention.

Unless otherwise stated, translational positions and velocities are expressed in the inertial frame, whereas angular velocities and body torques are expressed in the body frame. The Euler angles follow the $Z$--$Y$--$X$ sequence of $\mathbf R$ given above. Because this parameterization is singular at $\theta=\pm\pi/2$, the attitude representation and the operating envelope should be interpreted within the range where it remains valid; the termination guard $|\phi|,|\theta|\le1.2$~rad (Section~\ref{sec:metrics}) keeps all episodes well inside it.

Disturbances are modeled as force profiles rather than calibrated wind fields; equivalently, the wind velocity in the drag term $-A(\dot{\boldsymbol\xi}-\mathbf v_w)$ is set to zero. Blade flapping, ground effect, battery-voltage variation, and higher-order aerodynamics are omitted. The gyroscopic term vanishes to first order at balanced hover, so the linear predictors in Section~\ref{sec:ctrl} preserve the hover dynamics but neglect gyroscopic coupling away from hover.

\subsection{Rotor Convention, Allocation, and Motor Dynamics}\label{sec:rotor}
Rotors 1--4 are located at $(l,0),(0,-l),(-l,0),(0,l)$; rotors 1 and 3 spin counter-clockwise and rotors 2 and 4 clockwise, viewed from above. With rotor thrust $f_i=k\omega_i^2$ and generalized input $u=[T,\tau_\phi,\tau_\theta,\tau_\psi]^\top$,
\begin{equation}u=Mf,\quad M=\begin{bmatrix}1&1&1&1\\0&-l&0&l\\-l&0&l&0\\-b/k&b/k&-b/k&b/k\end{bmatrix}.\end{equation}
The signs of the roll, pitch, and yaw moments follow the rotor numbering, arm positions, and spin directions specified above. The same allocation matrix is used in the controller-to-rotor mapping, the plant model, the optimizer, and the verification scripts, where it is checked by applying collective thrust and single-axis torque commands to each rotor independently. The common allocator computes $f_{\rm raw}=M_0^{-1}u_{\rm req}$ from nominal parameters, clips each thrust to $[0,k_0\omega_{\max}^2]$, and converts thrusts to speed commands. Throughout, the subscript $0$ (as in $m_0$, $k_0$, $\mathbf I_0$, $A_0$, and $M_0$) denotes the nominal value used by the controllers and the allocator, whereas unsubscripted symbols denote the possibly perturbed plant value. The allocator does not prioritize thrust over torque under saturation, unlike prioritizing allocation schemes~\cite{faessler2017}. Three quantities are therefore distinguished. The \emph{requested} generalized input $u_{\rm req}$ is mapped to raw rotor thrusts with the nominal allocation matrix; each rotor-thrust command is then clipped independently before conversion to a rotor-speed command, so the \emph{allocated} generalized input may differ from the requested one when one or more rotors saturate. The \emph{applied} thrust and torque are determined by the actual rotor states and are not assumed to equal the requested input under saturation. This distinction underlies the applied-input variation metric of Section~\ref{sec:metrics}. Physical rotor speeds obey
\begin{equation}\tau_m\dot\omega_i=\omega_{c,i}-\omega_i,\qquad0\le\omega_i\le\omega_{\max}.\end{equation}
Here, $\tau_m$ is the motor time constant. The commanded speed $\omega_{c,i}$ is generated by the common allocator, whereas the physical speed $\omega_i$ evolves according to the motor dynamics and is constrained by the plant's rotor-speed limits.
The full plant thus has twelve rigid-body and four rotor states. Parametric uncertainty changes the plant coefficients while the controllers and allocator keep their nominal values. An optional motor-command slew limit is disabled in all primary experiments, and no additional virtual thrust or torque clipping is applied.

\emph{Reduced-authority case (RA).} To stress actuator authority, the RA case lowers the rotor-speed ceiling of the plant, allocator, and optimizer from 1000 to 650~rad/s. The maximum collective thrust then drops from $2.60\,m_0g$ to $4k_0\omega_{\max}^2=5.036$~N $\approx1.097\,m_0g$; the hover speed $\omega_h=\sqrt{m_0g/(4k_0)}\approx620.6$~rad/s is then 95\% of the ceiling. RA is therefore a severe stress test rather than a nominal operating condition.

\begin{table}[!t]\caption{Physical and numerical parameters. Physical coefficients follow~\cite{luukkonen2011}; actuator lag and speed ceiling are assumed.}\label{tab:parameters}\centering\footnotesize
	\begin{tabular}{lll}\toprule Symbol & Value & Unit\\\midrule
		$m$ & 0.468 & kg\\$l$ & 0.225 & m\\$g$ & 9.81 & m/s$^2$\\
		$I_{xx},I_{yy}$ & $4.856\times10^{-3}$ & kg\,m$^2$\\$I_{zz}$ & $8.801\times10^{-3}$ & kg\,m$^2$\\
		$k$ & $2.98\times10^{-6}$ & N\,s$^2$\\$b$ & $1.14\times10^{-7}$ & N\,m\,s$^2$\\
		$I_M$ & $3.357\times10^{-5}$ & kg\,m$^2$\\$\diag(A)$ & 0.25, 0.25, 0.25 & kg/s\\
		$\tau_m$ & 0.030 & s\\$\omega_{\max}$ (nominal / RA) & 1000 / 650 & rad/s\\
		Control period $T_s$ & 0.020 & s\\RK4 substep & 0.005 & s\\\bottomrule\end{tabular}\end{table}

\subsection{Simulation Interface}
\begin{figure}[!t]\centering
	\begin{tikzpicture}[font=\scriptsize, node distance=5mm and 2.6mm,
		blk/.style={draw, rounded corners=1pt, align=center, minimum height=8mm, inner sep=2pt},
		arr/.style={-{Latex[length=1.6mm]}}]
		\node[blk] (ref) {Reference\\(current/\\preview)};
		\node[blk, right=of ref] (ctrl) {Controller\\PID/LQR/\\SMC/MPC};
		\node[blk, right=of ctrl] (fb) {QP accept\\or LQR\\fallback$^\dagger$};
		\node[blk, right=of fb] (arr) {Command\\arrival\\(replay)$^\ddagger$};
		\node[blk, below=of arr] (alloc) {Common\\allocator\\+ clipping};
		\node[blk, left=of alloc] (mot) {Motor lag\\$\tau_m$, $\omega_{\max}$};
		\node[blk, left=of mot, minimum width=2.4cm] (plant) {Nonlinear rigid body\\RK4, $\mathbf F_d$,\\uncertainty};
		\draw[arr] (ref) -- (ctrl);
		\draw[arr] (ctrl) -- (fb);
		\draw[arr] (fb) -- (arr);
		\draw[arr] (arr) -- (alloc);
		\draw[arr] (alloc) -- (mot);
		\draw[arr] (mot) -- (plant);
		\draw[arr] (plant.west) -- ++(-3mm,0) |- node[pos=0.25, left]{$\mathbf X,\boldsymbol\omega$} (ref.west);
	\end{tikzpicture}
	\caption{Shared evaluation pipeline. All controllers use the same plant, allocator, motor model, and 50~Hz clock. $^\dagger$LMPC-F only. $^\ddagger$Active only in the timing replay of Section~\ref{sec:timing}.}\label{fig:arch}
\end{figure}
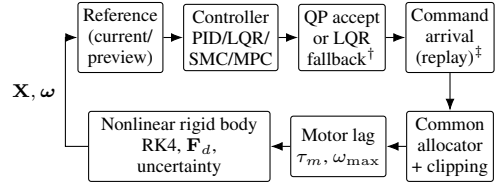

All controllers update at $T_s=20$~ms (50~Hz), so feedback, optimization, fallback selection, and command-arrival replay share one clock; 5~ms fourth-order Runge--Kutta (RK4) substeps integrate the plant between updates. Practical flight stacks typically run the attitude or body-rate loop at several hundred hertz and the position loop at a lower rate~\cite{faessler2017}. The single 50~Hz interface is a deliberate choice that gives every method the same timing budget and makes the measured MPC computation times directly comparable with the sampling period. Measurements are ideal and undelayed; only the timing experiment in Section~\ref{sec:timing} introduces command-arrival delays. The hover-model matrices $\mathsf A_c,\mathsf B_c$ used below are distinct from the drag matrix $A$. Fig.~\ref{fig:arch} summarizes the shared pipeline.

\section{Controller Implementations}\label{sec:ctrl}
All implementations receive ideal position, velocity, Euler angles, and body rates. They differ not only in feedback law but also in reference construction. PID and BL-SMC use the current position, velocity, and acceleration references, compute the desired attitude through a nonlinear force-to-attitude map, and compensate drag with the \emph{measured} velocity, $A_0\dot{\boldsymbol\xi}/m_0$. LQR and LMPC-F use small-angle hover references and compensate drag with the \emph{reference} velocity, $A_0\dot{\boldsymbol\xi}_r/m_0$. These architectural differences are part of each complete implementation and influence disturbance rejection alongside the feedback law and tuning. For brevity, tables use SMC for BL-SMC and MPC for LMPC-F.

\subsection{Cascade PID}
With $e_p=\boldsymbol\xi-\boldsymbol\xi_r$ and $e_v=\dot{\boldsymbol\xi}-\dot{\boldsymbol\xi}_r$, the commanded acceleration is
\begin{equation}
	a_c=a_r-K_pe_p-K_d\bar e_v-K_i\zeta+A_0\dot{\boldsymbol\xi}/m_0,
\end{equation}
where $\bar e_v[k]=\alpha\bar e_v[k-1]+(1-\alpha)e_v[k]$, $\alpha=e^{-T_s/0.03}$, is a filtered velocity error, so no position differentiation is required. The integrator $\zeta[k+1]=\operatorname{clip}(\zeta[k]+T_se_p[k],\pm2)$ is frozen when amplitude or slew limiting occurred at the previous sample (conditional-integration anti-windup).

Let $F_c=m_0(a_c+g\mathbf e_3)$. At zero yaw, the desired pitch is $\operatorname{atan2}(F_{c,x},\max(F_{c,z},0.5\,\mathrm N))$ and the desired roll is $\operatorname{atan2}(-F_{c,y},\sqrt{F_{c,x}^2+F_{c,z}^2})$, both clipped to $\pm0.55$~rad. The thrust is $T=\max(0,F_c^\top\mathbf R\mathbf e_3)$, and the wrapped attitude error $e_\eta$ drives
\begin{equation}
	\boldsymbol\tau=\mathbf I_0(-K_\eta e_\eta-K_\nu\boldsymbol\nu)+\boldsymbol\nu\times \mathbf I_0\boldsymbol\nu.
\end{equation}
The fixed gain diagonals are $K_p=(2.5,2.5,4)$, $K_d=(3,3,3.5)$, $K_i=(0.4,0.4,1)$, $K_\eta=(64,64,36)$, and $K_\nu=(16,16,12)$. Desired angular-rate feedforward is omitted to avoid derivative impulses on steps.

\subsection{Hover LQR}
At $u_0=[m_0g,0,0,0]^\top$ with $\delta u=u-u_0$, the hover model is
\begin{equation}
	\mathsf A_c=\begin{bmatrix}0&\mathbf I_3&0&0\\0&-A_0/m_0&G&0\\0&0&0&\mathbf I_3\\0&0&0&0\end{bmatrix},\
	\mathsf B_c=\begin{bmatrix}0&0\\\mathbf e_3/m_0&0\\0&0\\0&\mathbf I_0^{-1}\end{bmatrix},
\end{equation}
with $G=\left[\begin{smallmatrix}0&g&0\\-g&0&0\\0&0&0\end{smallmatrix}\right]$. State blocks have dimension three; input blocks have dimensions one and three. Exact zero-order-hold (ZOH) discretization gives $(\mathsf A_d,\mathsf B_d)$, and the solution $P$ of the discrete algebraic Riccati equation (DARE) gives
\begin{equation}
	K=(R_u+\mathsf B_d^\top P\mathsf B_d)^{-1}\mathsf B_d^\top P\mathsf A_d.
\end{equation}
The fixed weights are $\diag(Q)=(12,12,20,5,5,8,20,20,10,1,1,1)$ and $\diag(R_u)=(0.3,3,3,3)$ in SI units. With $a_f=a_r+A_0\dot{\boldsymbol\xi}_r/m_0$, the reference roll and pitch are $-a_{f,y}/g$ and $a_{f,x}/g$ with zero yaw and body rates, the feedforward is $u_{ff}=[m_0(g+a_{f,z}),0,0,0]^\top$, and $u=u_{ff}-K(\mathbf X-\mathbf X_r)$. This low-order reference is intended for moderate maneuvers. The reported LQR results characterize this fixed-gain controller based on the hover-linearized model and small-angle reference construction; they should not be interpreted as a general performance limit of LQR, particularly for aggressive maneuvers or designs that incorporate trajectory-dependent linearization and angular-rate feedforward.

\subsection{Boundary-Layer Sliding-Mode Control}
The position sliding variable and commanded acceleration are
\begin{align}
	s_p&=e_v+\Lambda_pe_p,\\
	a_c&=a_r-\Lambda_pe_v-K_s\sat(s_p/\varphi_p)+A_0\dot{\boldsymbol\xi}/m_0,
\end{align}
where $\sat(\cdot)$ is the componentwise unit saturation that replaces the sign function inside a boundary layer~\cite{slotine1991}. The same force-to-attitude map as PID is used. The attitude loop uses the local surrogate
\begin{align}
	s_\eta&=\boldsymbol\nu+\Lambda_\eta e_\eta,\\
	\boldsymbol\tau&=\mathbf I_0[-\Lambda_\eta\boldsymbol\nu-K_{s\eta}\sat(s_\eta/\varphi_\eta)]+\boldsymbol\nu\times \mathbf I_0\boldsymbol\nu,
\end{align}
which approximates Euler-error rates by body rates and omits desired-rate derivatives; it is therefore not an exact sliding variable on $SO(3)$. The fixed parameters are $\Lambda_p=(1.8,1.8,2.5)$~s$^{-1}$, $K_s=(1.2,1.2,1.8)$~m/s$^2$, $\varphi_p=(0.3,0.3,0.25)$~m/s, $\Lambda_\eta=(6,6,4)$~s$^{-1}$, $K_{s\eta}=(5,5,3)$~rad/s$^2$, and $\varphi_\eta=0.5$~rad/s per axis. For an ideal scalar channel with switching gain $\kappa$ and matched disturbance $|d|<\kappa$, the Lyapunov function $\mathcal V=s^2/2$ yields $\dot{\mathcal V}\leq-(\kappa-|d|)|s|$ outside the layer; the applied-input variation and motor traces assess the implemented behavior under sampling, coupling, lag, uncertainty, and clipping. The boundary-layer saturation reduces discontinuous switching but does not, by itself, establish stability of the complete sampled-data nonlinear system with actuator saturation, motor lag, and model uncertainty. The performance claims in this study therefore rest on the specified implementation and the reported simulation outcomes, not on a general stability guarantee.

\subsection{Linear MPC with Numerical Acceptance}
With $u_0=[m_0g,0,0,0]^\top$, LMPC-F optimizes $N=25$ moves (0.5~s):
\begin{align}
	\mathcal J={}&\sum_{j=0}^{N-1}\left[\norm{z_j-z_{r,j}}_Q^2+\norm{\delta u_j-\delta u_{ff,j}}_{R_u}^2\right. \left.+\norm{\delta u_j-\delta u_{j-1}}_S^2\right]+\norm{z_N-z_{r,N}}_P^2.\label{eq:mpc}
\end{align}
The rigid predictor uses $z=\mathbf X\in\mathbb R^{12}$, $z_0=\mathbf X[k]$, and $z_{j+1}=\mathsf A_dz_j+\mathsf B_d\delta u_j$. The fixed $Q,R_u$ equal the LQR weights, and $S=\diag(0.2,0.8,0.8,0.8)$; tuned LQR and MPC weights are selected independently. All variants impose rotor-thrust bounds $0\le M_0^{-1}(u_0+\delta u_j)\le f_{\max}\mathbf1$ and increment bounds $|\delta u_j-\delta u_{j-1}|\le d_u$, $d_u=[0.5,0.03,0.03,0.02]^\top$ (N, N\,m) per sample, with $f_{\max}=2.980$~N nominally and $1.259$~N in RA. Here $\delta u_{-1}$ is the generalized request issued at the previous sample, not the motor-applied input: the first accepted QP move or, after a fallback, the fallback request itself. For the raw fallback this is the unclipped LQR request, which may lie outside the rotor-thrust bounds.

The QP is solved by OSQP~\cite{stellato2020} with absolute and relative tolerances $10^{-5}$, at most 4000 iterations, adaptive-$\rho$ interval 50, warm starting, and no polishing. A solution is \emph{accepted} only if OSQP reports \textit{solved} or \textit{solved inaccurate}, all horizon values are finite, and the constraint excess is at most $10^{-3}$~N for rotor and collective increments, $10^{-4}$~N\,m for torque increments, and $10^{-4}$~rad for tilt constraints. Otherwise, the fallback is invoked. The increment penalty and bounds govern accepted QP actions only; the raw LQR fallback bypasses them, so the closed-loop LMPC-F signal can exceed the QP increment limits during recovery. All horizons, statuses, and acceptance decisions are logged.

A solver status of \textit{solved} is thus necessary but not sufficient for command acceptance. An accepted MPC solution provides a requested generalized input, which then passes through the common rotor allocator and the actuator model. Solver acceptance therefore does not imply that the request is reproduced exactly by the physical plant: saturation, motor lag, and command-arrival delays may make the applied input differ from the optimizer's request.

\subsection{Motor-State Prediction Variant}
Let $\omega_h=\sqrt{m_0g/(4k_0)}$ and $G_\omega=2k_0\omega_hM_0$. Augmenting the hover model with linearized rotor dynamics in the same generalized-request coordinates gives
\begin{equation}
	\begin{bmatrix}\dot{\mathbf X}\\\dot{\delta \boldsymbol\omega}\end{bmatrix}
	=\begin{bmatrix}\mathsf A_c&\mathsf B_cG_\omega\\0&-\mathbf I_4/\tau_m\end{bmatrix}\begin{bmatrix}\mathbf X\\\delta \boldsymbol\omega\end{bmatrix}
	+\begin{bmatrix}0\\G_\omega^{-1}/\tau_m\end{bmatrix}\delta u.
\end{equation}
Exact ZOH yields a 16-state predictor initialized with the measured $\boldsymbol\omega-\omega_h\mathbf1$. Its stage weight extends $Q$ with four zeros, its own DARE supplies the terminal weight $P$, and the augmented reference is $[\mathbf X_r^\top,\mathbf 0^\top]^\top$. The motor variant therefore changes three things at once: it adds measured rotor states, a motor-lag prediction model, and a model-derived terminal cost.

\subsection{Fallback and Tilt-Constraint Variants}
The \emph{raw} fallback applies the rigid LQR gain computed from the same $Q,R_u$ and bypasses the QP increment and tilt restrictions before the common allocator. The \emph{projected} fallback instead solves
\begin{equation}
	\begin{aligned}\min_{u_b}\;&\norm{L^{-1}(u_b-u_{\rm LQR})}^2\\
		\text{s.t.}\;&0\le M_0^{-1}u_b\le f_{\max}\mathbf1,\quad|u_b-u_{\rm prev}|\le d_u,\end{aligned}
\end{equation}
with $L=\diag(m_0g,m_0gl,m_0gl,m_0gl)$, using a separate four-variable OSQP instance with $10^{-8}$ tolerances; if the projection fails, the previous request is held and the event is logged. The projection restricts the input only; no invariant set, control-Lyapunov certificate, or recursive-feasibility argument is attached to it.

Three tilt-constraint placements are compared. \emph{Hard-initial} imposes $|\phi_j|,|\theta_j|\le0.45$~rad for $j=0,\ldots,N$; since $\mathbf X_0$ is fixed by equality, the QP is infeasible whenever the measured tilt already exceeds the bound. \emph{Hard-future} imposes the bound for $j=1,\ldots,N$ only. \emph{Soft-future} adds nonnegative slacks, $|a_j|\le0.45+\epsilon_{j,a}$, with a quadratic penalty $10^4\sum \epsilon_{j,a}^2$ and no upper slack bound. This penalty is not exact in the sense of~\cite{kerrigan2000}, so the tilt bound may be violated even when a feasible solution exists~\cite{kerrigan2000,zeilinger2014}. Unless stated otherwise, LMPC-F uses hard-initial tilt constraints and raw fallback. The \emph{preview-enabled} track (LMPC-F+P) is identical except that the known future reference schedule is supplied over the horizon.

\subsection{Geometric Baseline}
To include a controller designed on the rotation group, the confirmation study adds a geometric tracking controller in the spirit of~\cite{lee2010}. The commanded force $F_c=m_0(a_r-K_xe_p-K_ve_v+A_0\dot{\boldsymbol\xi}/m_0+g\mathbf e_3)$ defines the desired body axis $\mathbf b_{3d}=F_c/\norm{F_c}$, whose tilt is limited to 0.55~rad as for PID; with zero desired yaw, $\mathbf R_d=[\mathbf b_{2d}\times\mathbf b_{3d},\ \mathbf b_{2d},\ \mathbf b_{3d}]$ with $\mathbf b_{2d}=\mathbf b_{3d}\times\mathbf e_1/\norm{\mathbf b_{3d}\times\mathbf e_1}$. The thrust is $T=\max(0,F_c^\top\mathbf R\mathbf e_3)$ and
\begin{align}
	\boldsymbol\tau&=\mathbf I_0(-K_Re_R-K_\Omega\boldsymbol\nu)+\boldsymbol\nu\times\mathbf I_0\boldsymbol\nu,\\
	e_R&=\tfrac12(\mathbf R_d^\top\mathbf R-\mathbf R^\top\mathbf R_d)^\vee .
\end{align}
The fixed gains equal the PID position and attitude gains, $K_x=(2.5,2.5,4)$, $K_v=(3,3,3.5)$, $K_R=(64,64,36)$, $K_\Omega=(16,16,12)$. The desired angular velocity is set to zero and no integral action is used, so the controller differs from PID mainly in its attitude-error definition and its lack of integral action.

\section{Experimental Design}\label{sec:design}
\subsection{Tuning Protocol and Data Separation}
The execution protocol was fixed and time-stamped before the reported runs. Each controller receives twelve candidates: its fixed baseline plus eleven Latin-hypercube points~\cite{mckay1979} in a four-dimensional space of multipliers spanning 0.6--1.8. PID varies the position P, D, and I scales and an attitude-gain scale; LQR varies the position, velocity, and attitude $Q$ scales and the $R_u$ scale; BL-SMC varies the position surface, switching gain, boundary width, and attitude switching gain, with the vertical switching gain spanning an absolute range of 1.5--4.5~m/s$^2$, wider than the multiplier range used for the other parameters (a sensitivity check is reported in Section~\ref{sec:scope}); MPC varies the position and velocity $Q$ scales, $R_u$, and $S$. Tuning MPC weights also changes the associated LQR fallback gain.

Each candidate is scored on two development episodes (a modified figure-eight and a smooth transition, each with one fixed uncertainty and disturbance realization). The three best mean development scores each receive one validation episode, and the lowest validation score selects the setting (ties broken by candidate index). The score is
\begin{equation}J_{\rm tune}=100\chi_{\rm term}+\mathrm{RMSE}/(1\;\mathrm m)+0.05\log(1+V/(1\;\mathrm s^{-1})),\end{equation}
where $\chi_{\rm term}\in\{0,1\}$ indicates termination and $V$ is the applied-input variation defined in Section~\ref{sec:metrics}. Selected settings were hashed before the evaluation runs. The fixed track retains the historical gains and weights listed in Section~\ref{sec:ctrl}; these were not individually optimized, so the fixed-to-tuned change also reflects the quality of each baseline. Table~\ref{tab:tuning} summarizes the matched budget.

\begin{table}[!t]\caption{Matched tuning budget. Each method has 12 candidate settings (two development episodes each) and three validation finalists. IDs are zero-based. Wall time sums concurrent worker time.}\label{tab:tuning}\centering\footnotesize
	\setlength{\tabcolsep}{4pt}\begin{tabular}{lrrrrr}\toprule
		Method & Settings & Dev.+val. & Selected ID & Val. score & Wall min\\\midrule
		PID & 12 & 24 + 3 & 11 & 0.187 & 1.2\\
		LQR & 12 & 24 + 3 & 8 & 0.225 & 1.1\\
		SMC & 12 & 24 + 3 & 10 & 0.119 & 1.2\\
		MPC & 12 & 24 + 3 & 3 & 0.354 & 2.2\\
		\bottomrule\end{tabular}\end{table}

\subsection{Evaluation Realizations}
Each episode lasts 20~s. Fifty joint realizations randomize the reference and the plant: mass $\times[0.8,1.25]$, each inertia axis $\times[0.75,1.25]$, and thrust coefficient $\times[0.8,1.15]$ relative to nominal. The reference is a phased figure-eight,
\begin{equation}\boldsymbol\xi_r=\begin{bmatrix}\rho_x\sin(\omega_{tr}t+\beta)\\\rho_y\sin(2\omega_{tr}t+0.7\beta)\\1+\rho_z\sin(\omega_{tr,z}t+0.3\beta)\end{bmatrix},\end{equation}
with $(\rho_x,\rho_y,\rho_z)\in[0.8,1.25]\times[0.45,0.8]\times[0.1,0.2]$~m, $\omega_{tr}\in[0.4,0.55]$~rad/s, $\omega_{tr,z}\in[0.3,0.4]$~rad/s, $\beta\in[0,1.2]$, and analytical derivatives. The disturbance is a force pulse on $[5.4,11.4)$~s equal to $[0.7,-0.45,0.25]^\top$~N scaled by a uniform $[0.8,1.2]$ factor, plus colored noise with standard deviations $(0.2,0.2,0.1)$~N and correlation time 0.25~s. Initial position and velocity match the reference, attitude and rates are zero, and the motors start at the nominal hover speed $\omega_{h}=\sqrt{m_0g/(4k_0)}$; on a perturbed plant this acts as an initial thrust imbalance, with initial thrust-to-weight ratios between 0.64 and 1.44 over the sampled mass and thrust-coefficient ranges. Each realization is shared by all four methods and both tracks, giving 400 paired evaluation episodes. Evaluation seeds are disjoint from the development and validation seeds, and the selection procedure never accesses evaluation scores. The evaluation realizations had, however, been inspected during earlier development of the benchmark; Section~\ref{sec:scope} discusses this point. In the main evaluation, both MPC tracks repeat the current reference over the horizon (no preview), matching the information available to the other controllers: like LQR, LMPC-F still receives the current reference acceleration through its feedforward term $\delta u_{ff}$, but no future reference values. Because preview is standard in trajectory-tracking MPC, the confirmation study also evaluates the preview-enabled LMPC-F+P.

\subsection{MPC Mechanism Experiments}
\emph{Matrix A} crosses five binary factors for LMPC-F with fixed weights: preview (current reference repeated vs.\ known future schedule), predictor (rigid vs.\ motor), fallback (raw vs.\ projected), authority (nominal vs.\ RA), and reference (abrupt vs.\ smooth), giving 32 cells. The abrupt reference steps from $[0,0,1]^\top$ to $[2,2,1.5]^\top$~m at 3~s; it is an intentional stress input, not a feasible planned flight. The smooth reference connects the same endpoints over $D_t=4$~s as $\boldsymbol\xi_r=\boldsymbol\xi_0+d\,q(\sigma)$ with $d=[2,2,0.5]^\top$~m, $\sigma=\operatorname{clip}((t-3)/D_t,0,1)$, and $q=10\sigma^3-15\sigma^4+6\sigma^5$, so velocity and acceleration vanish at both ends. Its bounds are $\|\dot{\boldsymbol\xi}_r\|\le1.347$~m/s, $\|\ddot{\boldsymbol\xi}_r\|\le1.037$~m/s$^2$, and $\|\boldsymbol\xi_r^{(3)}\|\le2.693$~m/s$^3$, placing it between the figure-eight and the step in difficulty. \emph{Matrix B} adds 16 cells that replace hard-initial with hard-future or soft-future tilt constraints for the rigid/raw configuration across both preview settings, authorities, and references; its hard-initial comparator is taken from Matrix A. These 48 cells are deterministic diagnostics, not independent uncertainty trials. In total there are $96+12+48+400=556$ primary episodes; six dedicated timing profiles and six arrival replays are additional.

\subsection{Confirmation Study on Fresh Realizations}\label{sec:fresh}
After the primary study, a confirmation protocol was fixed and hashed before execution. It draws 50 new realizations from the same distribution (new seeds; none was generated or inspected before the analysis reported here) and evaluates six methods: the four primary implementations, LMPC-F+P, and the geometric controller. Each method receives 48 candidates: for the four primary methods, the first twelve are exactly the primary-study candidates, followed by 36 new Latin-hypercube points; LMPC-F+P uses the LMPC-F candidates with preview enabled, and the geometric controller varies the scales of $K_x$, $K_v$, $K_R$, and $K_\Omega$. Applying the same development/validation rule to the first 12, 24, and 48 candidates yields nested budgets. Re-executing the primary tuning episodes in this protocol reproduced the non-timing metrics (RMSE, $V$, termination, and development score) of all 96 shared development episodes exactly and selected the same four settings; only the measured timing fields differ. Because the three finalists of the nested budgets overlap, only the distinct finalists are validated (5--7 per method), which gives 36 validation episodes. The study comprises 576 development, 36 validation, and 850 evaluation episodes (fixed settings plus every distinct selected setting per method), i.e., 1462 episodes in addition to the 556 primary episodes.

\subsection{Metrics, Termination, and Statistics}\label{sec:metrics}
Position RMSE is $\sqrt{n^{-1}\sum\norm{\boldsymbol\xi-\boldsymbol\xi_r}^2}$; step experiments report the post-step window $t\ge3$~s. Using $L$ above, the applied-input variation is $V=T_s\sum\norm{\Delta(L^{-1}u)/T_s}^2$ in s$^{-1}$, computed from the motor-applied inputs. $V$ is an actuation-smoothness metric, not electrical energy or ideal sliding-mode chattering. Requested, allocated, and applied signals are logged separately.

An episode terminates if the state becomes nonfinite, the position error exceeds 10~m, or $|\phi|$ or $|\theta|$ exceeds 1.2~rad. These guards detect divergence and departure from the Euler/hover-model regime; they are not operational safety limits. Completion is reported before accuracy, RMSE rankings use complete episodes only, and partial RMSE of terminated episodes is shown as a diagnostic.

The statistical unit is the joint realization, not a time sample or QP call. Means and paired differences use 10\,000 episode-level percentile bootstrap resamples~\cite{efron1993}, and failure proportions use exact Clopper--Pearson 95\% intervals~\cite{clopper1934}. The reported intervals quantify episode-level uncertainty and are descriptive rather than multiplicity-adjusted inferential guarantees. Comparisons across multiple controllers and experimental contrasts should therefore be interpreted in the context of the stated analysis plan and the number of comparisons performed.

\subsection{Measured Timing and Command-Arrival Replay}\label{sec:timing}
After all study runs had finished, six mechanism configurations were re-executed sequentially on one Windows desktop with a single numerical thread and default process priority. Each profile includes the first call (no warm-up). The measured command-call time includes reference and diagnostic computations but excludes setup, allocation, plant integration, and logging. State, input, status, and horizon traces were checked against the corresponding study episode.

The same six profiles were then repeated three times on a Raspberry Pi~5 (four Cortex-A76 cores at 2.4~GHz, Raspberry Pi OS based on Debian~13, Linux~6.12, aarch64) with Python~3.13.5 and the same NumPy, SciPy, and OSQP versions. The process was pinned to one core, run at raised priority (\texttt{nice -10}) with the performance frequency governor and one numerical thread; the SoC temperature stayed between 65 and 70~$^\circ$C and no throttling was reported. On this board we also tested a deadline-aware variant of the four nominal profiles in which the OSQP iteration limit was lowered from 4000 to 400 or 100. A call that stops at the limit fails the acceptance rule and is handled by the LQR fallback like any other rejected solve.

Each measured latency trace is then replayed exogenously. At each sample, a candidate command is computed from the sample-start state. If its measured finish time exceeds 20~ms, it is discarded and the previous request is held; otherwise, the plant integrates under the previous motor command until the measured arrival time and under the new command afterwards (RK4 substeps $\le5$~ms). Late candidates still update the internal controller and solver state, while the MPC preceding-input state is reset to the applied or held request. If the original profile terminated early, its duration also limits the replay window (Table~\ref{tab:latency}).

\section{Results}\label{sec:results}
\subsection{Effect of the Tuning Budget}
\begin{table*}[!t]\caption{Evaluation of complete controller implementations over 20~s (50 paired realizations per method and track). RMSE and $V$ use complete episodes. Failure intervals are exact (Clopper--Pearson) 95\%; RMSE intervals are episode-bootstrap 95\%. Zero observed failures does not imply zero risk.}\label{tab:heldout}\centering\footnotesize
	\begin{tabular}{llrrlll}\toprule
		Track & Method & Complete & Fail & Failure CI (\%) & Mean RMSE [95\% CI] (m) & Mean $V$ (s$^{-1}$)\\\midrule
		fixed & PID & 50/50 & 0/50 & [0.0, 7.1] & 0.372 [0.359, 0.387] & 0.140\\
		fixed & LQR & 50/50 & 0/50 & [0.0, 7.1] & 0.300 [0.285, 0.315] & 1.710\\
		fixed & SMC & 50/50 & 0/50 & [0.0, 7.1] & 1.105 [0.795, 1.453] & 0.474\\
		fixed & MPC & 50/50 & 0/50 & [0.0, 7.1] & 0.396 [0.380, 0.413] & 400.873\\
		tuned & PID & 50/50 & 0/50 & [0.0, 7.1] & 0.226 [0.218, 0.235] & 0.335\\
		tuned & LQR & 50/50 & 0/50 & [0.0, 7.1] & 0.238 [0.226, 0.250] & 1.588\\
		tuned & SMC & 50/50 & 0/50 & [0.0, 7.1] & 0.115 [0.107, 0.124] & 1.212\\
		tuned & MPC & 50/50 & 0/50 & [0.0, 7.1] & 0.377 [0.363, 0.391] & 363.405\\
		\bottomrule\end{tabular}\end{table*}

\begin{table}[!t]\caption{Tuned minus fixed paired differences (50 both-complete pairs per method; bootstrap 95\% CIs, unadjusted).}\label{tab:paired}\centering\footnotesize
	\setlength{\tabcolsep}{2.5pt}\scriptsize
	\begin{tabular}{lrlrl}\toprule
		Method & $\Delta$RMSE (m) & 95\% CI & $\Delta V$ (s$^{-1}$) & 95\% CI\\\midrule
		PID & $-0.146$ & [$-0.152$, $-0.141$] & 0.195 & [0.170, 0.224]\\
		LQR & $-0.062$ & [$-0.066$, $-0.059$] & $-0.122$ & [$-0.149$, $-0.099$]\\
		SMC & $-0.990$ & [$-1.331$, $-0.688$] & 0.738 & [0.634, 0.856]\\
		MPC & $-0.019$ & [$-0.026$, $-0.011$] & $-37.468$ & [$-99.309$, 30.538]\\
		\bottomrule\end{tabular}\end{table}

\begin{figure*}[!t]\centering\includegraphics[width=.95\textwidth]{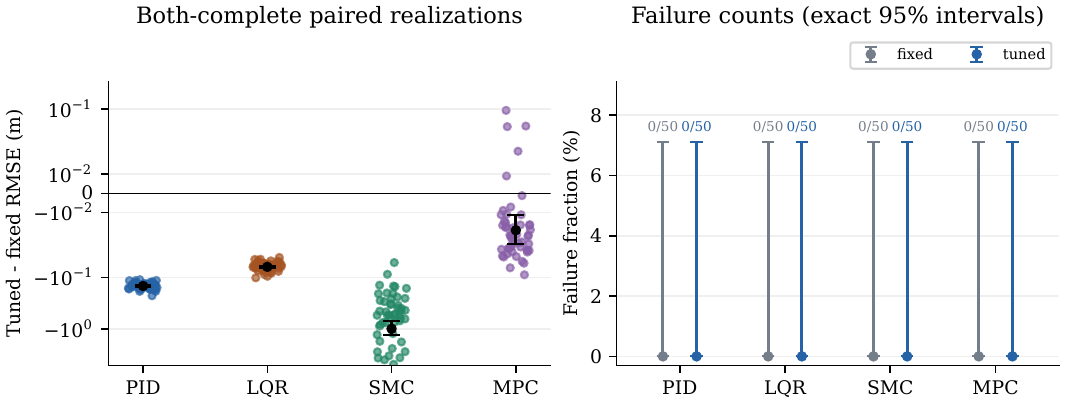}\caption{Left: tuned-minus-fixed RMSE differences for each realization (dots), with paired mean and bootstrap 95\% interval (black); the symmetric-log axis preserves the SMC tail. Right: failure counts out of 50 for every method and track, with exact 95\% binomial intervals.}\label{fig:paired}\end{figure*}

All eight method--track combinations complete all 50 evaluation realizations (Table~\ref{tab:heldout}). With the fixed settings, mean RMSE increases in the order LQR, PID, MPC, SMC; after tuning, the order becomes SMC, PID, LQR, MPC. Every paired difference is negative and every interval excludes zero (Table~\ref{tab:paired}, Fig.~\ref{fig:paired}). The largest reduction, 0.990~m, occurs for BL-SMC, whose fixed baseline has a highly uncertain mean (95\% CI [0.795, 1.453]~m), reflecting a few large-error episodes; the smallest, 0.019~m, occurs for LMPC-F. Paired contrasts relative to tuned PID support the primary-study ordering: BL-SMC differs by $-0.111$~m [$-0.116$, $-0.105$], LQR by $+0.012$~m [$0.003$, $0.020$], and LMPC-F by $+0.151$~m [$0.137$, $0.165$] (paired bootstrap 95\% intervals). A matched twelve-candidate budget is thus sufficient to change the mean-RMSE ordering. Most of this change comes from the poor fixed BL-SMC baseline, which moves from last to first; among PID, LQR, and LMPC-F the change is a single adjacent swap of PID and LQR (0.012~m apart after tuning). The change shows that rankings obtained from untuned or unequally tuned baseline settings can be strongly influenced by the tuning protocol and by the quality of the chosen baseline settings.

Accuracy gains do not imply smoother actuation. PID and BL-SMC reduce RMSE while increasing $V$, LQR reduces both, and the MPC change in $V$ has an interval that includes zero. With zero failures in 50 trials, the exact upper bound on the failure probability is still about 7.1\% per group.

\emph{Why LMPC-F has large input variation.} Tuned LMPC-F has $V=363.4$~s$^{-1}$, versus 1.588~s$^{-1}$ for LQR with which it shares its model and fallback gain. Over the 50 tuned episodes, the fallback is invoked on 4793 of 50\,000 calls (9.59\%): 3743 primal-infeasibility statuses, 925 iteration-limit statuses, and 125 solved statuses rejected by the acceptance check. Only 111 calls start outside the initial tilt envelope, so the hard-initial tilt constraint alone cannot explain the fallback frequency; the logs suggest that infeasibility arises mainly from the interaction of the increment bounds $d_u$, the rotor-thrust bounds, and the disturbance. A retrospective decomposition attributes 92.4\% of total $V$ to increments in which at least one endpoint is a fallback action, but only 12.3\% to the 1440 QP/fallback mode transitions (28.8 per episode). Most of the variation therefore arises \emph{within} fallback intervals, where the LQR gain is applied without increment limits and its output is shaped only by allocator clipping and motor lag, rather than solely at mode switches. Consequently, penalizing and bounding QP increments does not bound the variation of the complete hybrid controller. Because removing the fallback would change the trajectories themselves, this decomposition is descriptive rather than counterfactual. One possible contributor follows from the definition of $\delta u_{-1}$ in Section~\ref{sec:ctrl}: after a raw fallback, the next QP must start its increment bound from the unclipped LQR request, and if no input within $d_u$ of that request satisfies the rotor-thrust bounds, the next QP is infeasible as well, so one fallback can lead to a run of further fallbacks. The present logs do not separate this mechanism from infeasibility caused by the disturbance, and its share of the 3743 infeasible calls has not been quantified.

\subsection{Confirmation on Fresh Realizations and Larger Budgets}
\begin{table*}[!t]\caption{Confirmation study on 50 fresh realizations (Section~\ref{sec:fresh}). Mean RMSE over complete episodes with bootstrap 95\% CI; ``Fail'' counts terminations out of 50. Budget columns use the setting selected from the first 12, 24, or 48 candidates; identical selections give identical values. $V$ and fallback percentage refer to the 12-candidate selection. BL-SMC values use the wider vertical-switching-gain range of Section~\ref{sec:design}; with the common multiplier range, tuned BL-SMC reaches 0.228~m (Section~\ref{sec:scope}).}\label{tab:fresh}\centering\footnotesize
	\setlength{\tabcolsep}{2.5pt}
    \begin{tabular}{@{}lllllrr@{}}\toprule
		Method & Fixed: RMSE [CI] (m), Fail & Budget 12 (m) & Budget 24 (m) & Budget 48 (m) & $V_{12}$ (s$^{-1}$) & FB$_{12}$ (\%)\\\midrule
		PID & 0.385 [0.363, 0.408], 0 & 0.231 [0.219, 0.242] & 0.231 [0.219, 0.242] & 0.223 [0.212, 0.234] & 0.332 & --\\
		LQR & 0.295 [0.283, 0.307], 0 & 0.235 [0.225, 0.246] & 0.235 [0.225, 0.246] & 0.235 [0.225, 0.246] & 1.530 & --\\
		BL-SMC & 1.096 [0.791, 1.437], 2 & 0.128 [0.113, 0.146] & 0.128 [0.113, 0.146] & 0.128 [0.113, 0.146] & 1.175 & --\\
		LMPC-F & 0.385 [0.371, 0.399], 0 & 0.370 [0.356, 0.383] & 0.352 [0.339, 0.364] & 0.351 [0.340, 0.363] & 193.6 & 5.7\\
		LMPC-F+P & 0.301 [0.289, 0.313], 0 & 0.268 [0.257, 0.280] & 0.249 [0.238, 0.259] & 0.249 [0.238, 0.259] & 1.299 & 0.0\\
		Geometric & 0.544 [0.499, 0.593], 0 & 0.341 [0.314, 0.369] & 0.345 [0.317, 0.376] & 0.345 [0.317, 0.376] & 0.171 & --\\
		\bottomrule\end{tabular}\end{table*}

\begin{figure*}[!t]\centering\includegraphics[width=.9\textwidth]{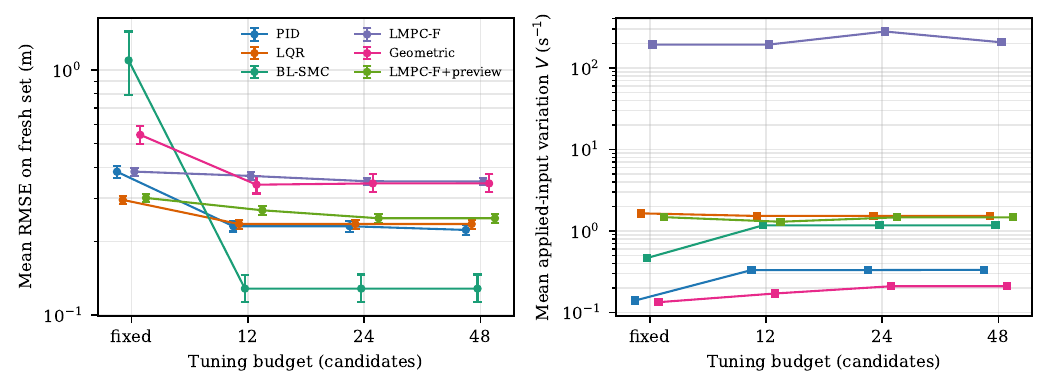}\caption{Confirmation study on 50 fresh realizations: mean RMSE with bootstrap 95\% intervals (left) and mean applied-input variation (right) for fixed settings and for the settings selected with budgets of 12, 24, and 48 candidates. Both axes are logarithmic.}\label{fig:budget}\end{figure*}

\emph{Reproducibility of the ranking.} On the fresh realizations, the tuned errors of the four primary methods are within 0.013~m of their primary-study values (PID 0.231 vs.\ 0.226, LQR 0.235 vs.\ 0.238, BL-SMC 0.128 vs.\ 0.115, LMPC-F 0.370 vs.\ 0.377~m), and the fixed-to-tuned change in the ordering recurs (Table~\ref{tab:fresh}); fixed PID and fixed LMPC-F are nearly tied on the fresh set (0.3847 vs.\ 0.3852~m). If the BL-SMC vertical switching gain is restricted to the common multiplier range, however, tuned BL-SMC is no longer distinguishable from tuned PID on the fresh set (0.228 vs.\ 0.231~m; paired difference $-0.003$~m [$-0.100$, $0.140$]; Section~\ref{sec:scope}). The first place of BL-SMC on the fresh set therefore depends on that search range. The fixed BL-SMC now fails in 2 of 50 episodes (position guard at 12.6 and 14.1~s; exact 95\% CI 0.5--13.7\%), whereas it completed all 50 primary realizations. This is consistent with the 7.1\% upper bound reported in Table~\ref{tab:heldout} and illustrates why zero observed failures must not be read as zero risk.

\emph{Effect of a larger budget.} Doubling or quadrupling the budget changed the selected setting for PID (48 only), LMPC-F, LMPC-F+P, and the geometric controller, but the paired changes relative to the 12-candidate selection were small: $-0.008$~m for PID, $-0.018$ and $-0.019$~m for LMPC-F, $-0.019$~m for LMPC-F+P, and $+0.004$~m for the geometric controller, whose larger-budget choice won on the single validation episode but generalized slightly worse. The mean-RMSE ordering was identical for budgets of 12, 24, and 48 (Fig.~\ref{fig:budget}). Not every adjacent pair in this ordering is statistically distinguishable on the fresh set: tuned PID and LQR differ by $-0.005$~m [$-0.013$, $0.003$] at budgets 12 and 24 and by $-0.013$~m [$-0.020$, $-0.006$] at 48, and the geometric controller and LMPC-F overlap at budgets 24 and 48 (paired bootstrap 95\% intervals). For the tested task and candidate-generation procedure, the mean-RMSE ordering remained unchanged as the tuning budget increased from 12 to 48 candidates. This supports the robustness of the observed ordering change within the evaluated search range, but it does not establish budget-independent rankings for other tasks or parameterizations.

\emph{Preview and the fallback tail.} Without preview, LMPC-F invoked its fallback in only 8 (fixed) and 10 (tuned) of the 50 fresh episodes, yet these episodes dominate its input variation: the median $V$ is 1.7~s$^{-1}$, comparable to LQR, while the five largest episodes account for 72--79\% of the total. With preview enabled, no fallback occurred in any of the 150 LMPC-F+P episodes (fixed setting and both tuned selections), the maximum $V$ was 3.1~s$^{-1}$, and the paired differences relative to LMPC-F were $-0.102$~m [$-0.111$, $-0.093$] in RMSE and $-192$~s$^{-1}$ [$-326$, $-76$] in $V$. The extreme mean variation of LMPC-F in Table~\ref{tab:heldout} is thus a heavy-tailed effect of a few fallback episodes, and in this task such episodes occurred only when the controller planned toward a stationary (repeated) reference. Even with preview, LMPC-F+P remains less accurate than tuned PID, LQR, and BL-SMC. Since tuned LQR uses the same hover-based reference and reference-velocity drag feedforward, this gap is specific to the MPC formulation; plausible contributors are the increment bounds $d_u$, which limit how quickly the input can respond to the force pulse, and the horizon cost structure.

\emph{Geometric baseline.} After tuning, the geometric controller reaches 0.341~m, i.e., 0.110~m [0.091, 0.131] worse than PID, with the lowest input variation of all methods. Since it shares PID's force-to-attitude structure but lacks integral action, the difference is consistent with the uncompensated 6-s force pulse; it should not be interpreted as a limitation of geometric control in general.

\subsection{Preview, Motor Prediction, and Fallback Interaction}
\begin{table*}[!t]\caption{Abrupt-step factorial (Matrix A, fixed weights, hard-initial tilt). ``F $t$'' denotes termination at $t$~s; starred RMSE values are partial post-step diagnostics excluded from ranking. FB: percentage of fallback calls. Pred. error: mean one-step position-prediction error evaluated along each configuration's own trajectory. Because trajectories and operating states differ across configurations, these values are descriptive and should not be interpreted as a matched-state comparison of predictor accuracy.}\label{tab:ablation}\centering\footnotesize
	\begin{tabular}{llllllll}\toprule
		Case & Preview & Predictor & Fallback & Outcome & Post RMSE (m) & FB (\%) & Pred. error (m)\\\midrule
		RA & no & motor & projected & F 3.44 & 2.837$^*$ & 11.6 & 0.00005\\
		RA & no & motor & raw & F 4.32 & 2.521$^*$ & 29.6 & 0.00062\\
		RA & no & rigid & projected & F 3.46 & 2.830$^*$ & 7.5 & 0.00007\\
		RA & no & rigid & raw & F 4.62 & 2.277$^*$ & 30.7 & 0.00205\\
		RA & yes & motor & projected & complete & 0.512 & 1.4 & 0.00001\\
		RA & yes & motor & raw & F 4.36 & 2.387$^*$ & 34.4 & 0.00059\\
		RA & yes & rigid & projected & F 3.16 & 2.603$^*$ & 14.6 & 0.00007\\
		RA & yes & rigid & raw & F 3.36 & 2.787$^*$ & 19.6 & 0.00073\\
		nominal & no & motor & projected & F 3.30 & 2.857$^*$ & 7.3 & 0.00003\\
		nominal & no & motor & raw & complete & 0.568 & 2.1 & 0.00001\\
		nominal & no & rigid & projected & F 3.34 & 2.849$^*$ & 6.0 & 0.00005\\
		nominal & no & rigid & raw & complete & 0.650 & 3.5 & 0.00003\\
		nominal & yes & motor & projected & F 3.58 & 2.641$^*$ & 24.6 & 0.00012\\
		nominal & yes & motor & raw & complete & 0.584 & 5.6 & 0.00002\\
		nominal & yes & rigid & projected & F 3.10 & 2.508$^*$ & 14.8 & 0.00012\\
		nominal & yes & rigid & raw & complete & 0.574 & 5.4 & 0.00005\\
		\bottomrule\end{tabular}\end{table*}

\begin{figure*}[!t]\centering\includegraphics[width=.85\textwidth]{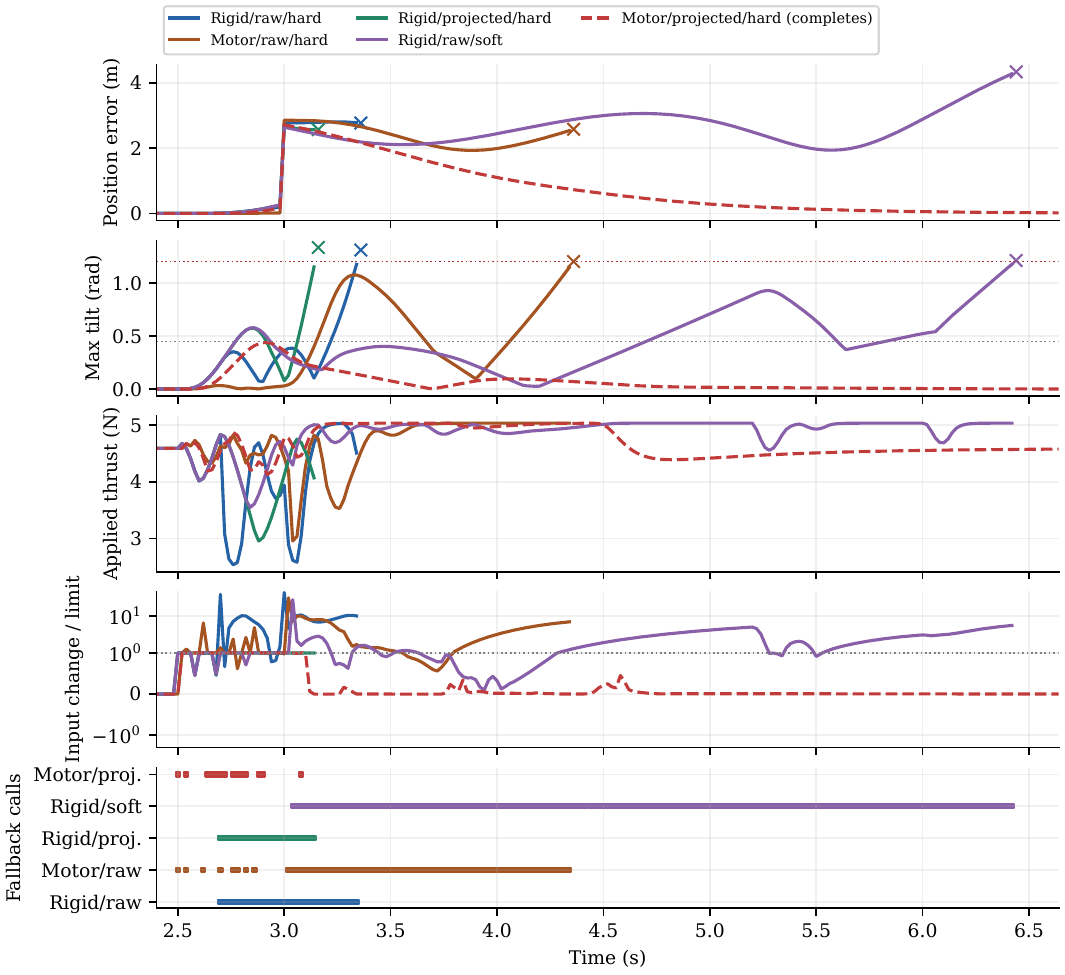}\caption{Abrupt step with scheduled preview in the reduced-authority (RA) case. The dashed trace (motor predictor, projected fallback) is the only RA configuration that completes. All variants share the allocator and plant; crosses mark terminations. The input-increment ratio is normalized by the componentwise QP limits $d_u$; bottom rows mark fallback calls. Projection restricts the requests but not the nonlinear state evolution.}\label{fig:hybrid}\end{figure*}

\emph{Smooth maneuver.} All sixteen smooth-reference cells complete. Under nominal authority, preview reduces post-maneuver RMSE from 0.16887 to 0.00295~m with the rigid predictor and from 0.17024 to 0.00312~m with the motor predictor. For the tested smooth maneuver, reference preview has a substantially larger effect on tracking error than the choice between the two evaluated predictors; without preview, the MPC repeatedly plans to a stationary target and lags the moving one. This result is specific to the tested reference and implementation; the abrupt-step experiments below show that preview interacts with fallback policy and actuator authority in a different way.

\emph{Abrupt step.} The pattern reverses for the step (Table~\ref{tab:ablation}). Raw fallback completes four of eight cells, projected fallback only one. Under nominal authority, every raw-fallback variant completes and every projected variant terminates. Under RA, every raw-fallback variant terminates, and the only completing cell combines motor prediction, preview, and projected fallback (post-step RMSE 0.512~m, 1.4\% fallback). Fig.~\ref{fig:hybrid} illustrates these trajectories for the RA case with preview. These counts do not indicate a universally superior fallback policy: under nominal authority raw fallback completes all four tested variants, whereas under reduced authority it terminates in all four, and only the preview-enabled motor-predictor configuration with projected fallback completes. Within these runs, the outcome of the fallback policy depends strongly on the remaining design choices. Each cell is a single deterministic trajectory, however, and Section~\ref{sec:timing-results} shows that floating-point differences between two machines alone moved the termination of one RA profile from 4.36 to 3.30~s. Near the stability boundary a binary outcome can therefore change under small perturbations, and the counts above describe the tested trajectories rather than completion rates.

The logged trajectories suggest the following mechanism, which the present design cannot isolate causally. After a 2.9~m step, the QP is frequently infeasible because the increment bounds prevent the input from changing quickly enough. The raw fallback then supplies large corrective requests that recover the nominal vehicle, but under RA these saturate the rotors and the tilt diverges. The projected fallback limits the requests to the increment bounds, the same bounds under which the QP has just failed; with ample authority this slows recovery until the tilt guard is reached, whereas with reduced authority and accurate motor prediction it avoids saturation-induced divergence. Because projection also alters subsequent states, optimizer acceptance, and allocation, the contrasts identify an interaction rather than a single causal path. The practical message is that the fallback policy is a first-order design choice whose best setting depends on the task and the actuator margin.

\subsection{Tilt-Constraint Placement and Relaxation}
\begin{table}[!t]\caption{RA abrupt step, rigid predictor, raw fallback (Matrix B). Hard-future omits only the $j=0$ tilt bound; soft-future allows penalized slack. Starred values are partial.}\label{tab:tilt}\centering\footnotesize
	\begin{tabular}{lllll}\toprule
		Preview & Tilt policy & Outcome & Post RMSE & Max slack (rad)\\\midrule
		no & hard-initial & F 4.62 & 2.277$^*$ & 0.00000\\
		no & hard-future & F 4.62 & 2.277$^*$ & 0.00000\\
		no & soft-future & F 4.62 & 2.278$^*$ & 0.00341\\
		yes & hard-initial & F 3.36 & 2.787$^*$ & 0.00000\\
		yes & hard-future & F 3.36 & 2.787$^*$ & 0.00000\\
		yes & soft-future & F 6.44 & 2.679$^*$ & 0.12700\\
		\bottomrule\end{tabular}\end{table}

For the RA step with preview, removing the initial tilt bound leaves the termination time unchanged at 3.36~s, whereas penalized future slack extends the trajectory to 6.44~s before termination (Table~\ref{tab:tilt}). An initial state outside a hard tilt envelope makes the QP infeasible, but this is not what limits recovery here: the outcome is governed by actuator authority, and softening the constraint only delays the failure. Soft-constrained solutions satisfy the relaxed rather than the original tilt bound (maximum slack 0.127~rad). The longer time to termination therefore indicates delayed failure under the relaxed formulation, not certified satisfaction of the original tilt limit.

\subsection{Sensitivity to Measured Command-Call Delays}\label{sec:timing-results}
\begin{table*}[!t]\caption{Dedicated timing profiles (first call included, no warm-up) and exogenous command-arrival replay on a Windows desktop (Intel Core i5-8400) and a Raspberry Pi~5 (one pinned core, median of three repeats; ranges are shown when repeats differ). Call time includes reference and diagnostic computations and excludes setup, allocation, integration, and logging. P99 is the 99th percentile of the call time. ``Window'' is the profile duration available for replay; a window shorter than 20~s cannot establish completion. Late/used: deadline misses among the replayed calls.}\label{tab:latency}\centering\footnotesize
\setlength{\tabcolsep}{2.4pt}\resizebox{\textwidth}{!}{\begin{tabular}{lll rrrll rrrll}\toprule
& & & \multicolumn{5}{c}{Desktop} & \multicolumn{5}{c}{Raspberry Pi 5}\\\cmidrule(lr){4-8}\cmidrule(lr){9-13}
Case & Prev. & Pred. & P99 (ms) & Miss (\%) & Window (s) & Replay & Late/used & P99 (ms) & Miss (\%) & Window (s) & Replay & Late/used\\\midrule
nominal & no & rigid & 101.8 & 8.9 & 20.00 & F 3.56 & 13/178 & 121.1 & 9.6 & 20.00 & F 3.58 & 14/179\\
nominal & yes & rigid & 100.1 & 6.9 & 20.00 & F 2.88 & 8/144 & 39.3 & 2.1 & 20.00 & F 2.88 & 9/144\\
nominal & no & motor & 32.2 & 1.5 & 20.00 & complete & 15/1000 & 37.4 & 1.6 & 20.00 & F 3.74--3.80 & 16--17/187--190\\
nominal & yes & motor & 95.1 & 5.9 & 20.00 & F 3.10 & 26/155 & 147.5 & 11.1 & 20.00 & F 3.60--3.80 & 35--41/180--190\\
RA & yes & rigid & 43.8 & 7.1 & 3.36 & no F in window & 12/168 & 53.6 & 7.1 & 3.36 & F 3.28 & 12/164\\
RA & yes & motor & 47.0 & 8.3 & 4.36 & F 4.12 & 18/206 & 79.9 & 12.1 & 3.30$^\dagger$ & no F in window & 19--20/165\\
\bottomrule\end{tabular}}

\smallskip
\parbox{\textwidth}{\footnotesize $^\dagger$Without delays, this profile terminates at 3.30~s on the Pi and at 4.36~s on the desktop, so the RA rows are not trajectory-matched across machines.}
\end{table*}

\begin{table*}[!t]\caption{Deadline-aware OSQP iteration caps on the Raspberry Pi~5 (nominal step, raw fallback, three repeats each; median values, maximum over repeats for ``Max''). Cap hits are calls out of 1000 that used the full iteration allowance; some of them still reached \textit{solved} or \textit{solved inaccurate} at the final convergence check and were accepted, so the fallback share can be smaller than the cap-hit share. RMSE is the post-step position RMSE.}\label{tab:caps}\centering\footnotesize
\setlength{\tabcolsep}{3pt}\resizebox{\textwidth}{!}{\begin{tabular}{rll rrrrr r ll}\toprule
Cap & Prev. & Pred. & P99 (ms) & Max (ms) & Miss (\%) & Cap hits & Fallback (\%) & No-delay RMSE (m) & Replay (3 runs) & Replay RMSE (m)\\\midrule
4000 & no & rigid & 121.1 & 140.6 & 9.6 & 14 & 3.5 & 0.650 & F 3.58 (3/3) & --\\
4000 & yes & rigid & 39.3 & 123.9 & 2.1 & 6 & 3.0 & 0.568 & F 2.88 (3/3) & --\\
4000 & no & motor & 37.4 & 191.6 & 1.6 & 4 & 2.0 & 0.568 & F 3.74--3.80 (3/3) & --\\
4000 & yes & motor & 147.5 & 221.8 & 11.1 & 9 & 11.5 & 0.606 & F 3.60--3.80 (3/3) & --\\\midrule
400 & no & rigid & 13.3 & 14.0 & 0.0 & 11 & 2.0 & 0.551 & complete (3/3) & 0.555--0.556\\
400 & yes & rigid & 13.6 & 14.3 & 0.0 & 34 & 4.0 & 0.537 & complete (3/3) & 0.542--0.550\\
400 & no & motor & 20.0 & 67.5 & 1.0 & 14 & 2.3 & 0.551 & complete (3/3) & 0.589--0.605\\
400 & yes & motor & 20.5 & 24.5 & 2.3 & 34 & 4.4 & 0.550 & complete 1/3; F 2.96, F 3.34 & 0.579\\\midrule
100 & no & rigid & 4.5 & 6.1 & 0.0 & 23 & 2.4 & 0.550 & complete (3/3) & 0.552--0.553\\
100 & yes & rigid & 4.5 & 4.9 & 0.0 & 51 & 4.8 & 0.550 & complete (3/3) & 0.551\\
100 & no & motor & 6.3 & 7.9 & 0.0 & 28 & 2.5 & 0.550 & complete (3/3) & 0.552--0.553\\
100 & yes & motor & 6.5 & 7.2 & 0.0 & 60 & 5.5 & 0.550 & complete (3/3) & 0.553\\
\bottomrule\end{tabular}}\end{table*}

\begin{figure}[!t]
	\centering
	\includegraphics[width=.5\columnwidth]{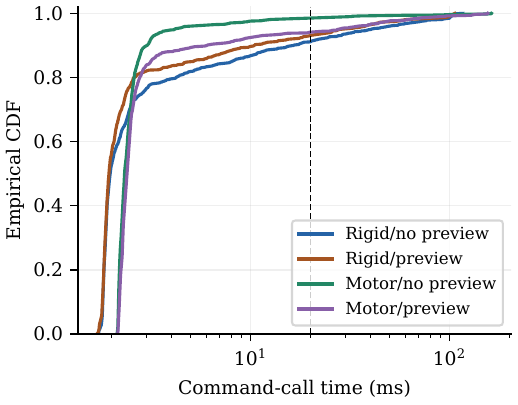}
	\caption{Empirical cumulative distribution function (CDF) of command-call times on the Windows desktop for the nominal step (1000 sequential calls per predictor/preview variant, raw fallback). The dashed line marks the 20~ms deadline. Calls are correlated; observed maxima are not worst-case execution-time bounds.}
	\label{fig:ecdf}
	\end{figure}

All four nominal raw-fallback step profiles complete 20~s with instantaneous command application, yet their 99th-percentile (P99) command-call times of 32.2--101.8~ms exceed the 20~ms period, although median calls are much shorter (Fig.~\ref{fig:ecdf}). Under measured-trace replay, both rigid-predictor variants and the motor/preview variant terminate, at 3.56, 2.88, and 3.10~s, respectively (Table~\ref{tab:latency}). Only the motor/no-preview variant completes, with post-step RMSE rising from 0.568 to 0.636~m. In the terminating rigid/no-preview replay, 13 of the 178 calls executed before termination miss the deadline (7.3\%), so a few percent of held commands during the post-step transient suffice to cause failure. A held request combined with delayed arrivals changes the plant state and therefore the next QP, compounding the effect. Numerical acceptance and timely delivery are therefore distinct properties of the same implementation.

\emph{Why the motor predictor is faster in the tail.} Without preview, the rigid and motor predictors both have a median of 25 OSQP iterations, but their 99th percentiles are 4000 and 725.5 iterations, with 15 and 3 calls reaching the iteration limit, respectively. The motor variant actually has a slightly higher median command time (2.328 vs.\ 1.970~ms) but a lower solver-only P99 (28.3 vs.\ 100.3~ms). The difference thus lies entirely in the convergence tail. Convergence of the alternating direction method of multipliers (ADMM) used by OSQP depends on residual behavior, constraint geometry, scaling, and warm starts rather than on state dimension alone~\cite{stellato2020}. A plausible explanation is that the motor model, with discrete pole $e^{-T_s/\tau_m}\approx0.51$, produces plans consistent with the actual rotor response, so warm starts from the shifted previous solution lie closer to the new optimum and fewer constraints change activity between calls. The logged final residuals are consistent with this hypothesis but cannot confirm it; conditioning and dual-oscillation effects remain to be analyzed.

Under RA, the rigid/preview replay shows no termination within its 3.36~s window, which does not establish 20~s completion, and the motor/preview replay terminates at 4.12~s, earlier than the 4.36~s of the original profile. The replay isolates the effect of the measured command-arrival traces under the stated simulation interface. It does not reproduce the scheduling, processor contention, communication, or actuator-interface effects of an embedded flight stack, so the observed failures demonstrate sensitivity to delayed command delivery in these profiles rather than a hardware-independent deadline-failure rate.

\emph{Embedded board.} Table~\ref{tab:latency} also lists the Raspberry Pi~5 results. The median call took 2.2--2.9~ms, 1.1--1.2 times the desktop value, and a solve that ran to 4000 iterations took about 119~ms, against 100~ms on the desktop. The board is therefore only slightly slower; what sets P99 is how many calls fall into the iteration tail. Across the three repeats of the uncapped profiles, P99 varied by less than 5\% and the replay outcomes were the same apart from small shifts in failure time. On the Pi, all four nominal profiles terminate under replay, including motor/no-preview, which completed on the desktop. Its miss rate is almost the same on both machines (1.6 vs.\ 1.5\%), but on the Pi 16--17 of the first 187--190 calls were late, so the misses fell in the post-step transient. P99 itself moved in both directions (39.3 vs.\ 100.1~ms for rigid/preview, 147.5 vs.\ 95.1~ms for motor/preview). Over 1000 calls, P99 is about the tenth-largest call, so it depends on whether a few solves reach the 4000-iteration limit, and those solves are sensitive to floating-point differences between the x86-64 and ARM builds of the same packages. The same sensitivity changes the no-delay RA motor/preview profile, which ends at 3.30~s on the Pi and at 4.36~s on the desktop, and the no-delay nominal preview profiles: post-step RMSE 0.568 vs.\ 0.574~m (rigid) and 0.606 vs.\ 0.584~m (motor), with fallback shares of 3.0 vs.\ 5.4\% and 11.5 vs.\ 5.6\% on the Pi and desktop (Tables~\ref{tab:ablation} and~\ref{tab:caps}); the no-preview profiles agree in post-step RMSE and differ by at most 0.1 percentage point in fallback share. The two machines also used different measurement protocols (desktop: single run at default priority with Python~3.12.14; Pi: median of three repeats on a pinned core at raised priority with Python~3.13.5), so cross-machine P99 comparisons are indicative.

\emph{Deadline-aware iteration caps.} Bounding the number of solver iterations is established practice in real-time MPC~\cite{wang2010}; the purpose here is to measure how much of the replay failure it removes on this benchmark. Table~\ref{tab:caps} shows the effect of tying the iteration limit to the deadline. With a cap of 400, P99 falls to 13--21~ms. Both rigid variants then miss no deadline and the motor/no-preview variant completes in all three replays, but the motor/preview variant, whose P99 is just above 20~ms, completes in only one of three. With a cap of 100, P99 is 4.5--6.5~ms, no call exceeds 8~ms, no deadline is missed, and all twelve replays complete with post-step RMSE within 0.003~m of the no-delay run. Between 2.3\% and 6.0\% of calls use the full iteration allowance; those that end with \textit{maximum iterations reached} or an infeasibility status are replaced by the fallback, and the total fallback share is 2.4--5.5\% of calls, compared with 2.0--11.5\% without the cap, and the no-delay post-step RMSE is 0.550~m in all four capped configurations versus 0.568--0.650~m uncapped. Calls of 64--68~ms in two of the cap-400 motor/no-preview repeats, where P99 stays at 20~ms, show that operating-system jitter remains even when the solver work is bounded. On this maneuver, the deadline failures therefore come from the unbounded iteration tail rather than from the processor class, and a cap with a clear margin to the deadline removes them with a slightly lower no-delay post-step RMSE; a plausible reason is an earlier hand-over to the LQR fallback during the post-step transient. This is one abrupt step on one board without a real-time kernel, so it does not show that capping is harmless in general.

\section{Discussion}\label{sec:discussion}
\subsection{Verification and Reproducibility}
Independently written test scripts check the physical coefficients, rotor mixing, hover equilibrium, gyroscopic coupling, and motor-hover Jacobian. A replay with zero arrival delay reproduces the reference implementation to within $10^{-12}$. An analytic ZOH audit re-evaluates all 152\,116 candidates with successful solver status: 151\,815 pass and 301 are rejected, with no disagreement with the online acceptance decisions. The audit uses dynamics and tilt tolerances of $10^{-4}$ per rigid-state block and $10^{-2}$~rad/s per motor state ($\approx7\times10^{-5}$~N at hover). A rerun of a copied source tree in a fresh offline environment with pinned dependencies reproduces the non-timing arrays of all 556 primary episodes and the six arrival replays within $10^{-12}$. This agreement holds for the same build; the Raspberry Pi runs in Section~\ref{sec:timing-results} show that x86-64 and ARM builds of the same packages can produce different trajectories near the stability boundary. The confirmation study was executed with the same pinned environment (Python 3.12.14, NumPy 2.5.3, SciPy 1.18.1, OSQP 1.1.3); four primary evaluation episodes (identical RMSE and $V$) and the non-timing metrics of all 96 shared development episodes were reproduced exactly before the new runs. The released material contains the configurations, horizons, failure records, selection results, statistical summaries, and timing traces of both studies.

\subsection{Practical Implications}
Table~\ref{tab:checklist} condenses the findings into a reporting checklist. Three implications stand out for practitioners. First, a ranking obtained from untuned or unequally tuned baseline settings can be strongly influenced by both tuning effort and baseline quality; in this study, a matched twelve-candidate search was sufficient to change the mean-RMSE ordering, the change recurred on fresh realizations, and a fourfold larger candidate-count budget did not change the ordering. The search space matters as well: restricting one BL-SMC gain to the common multiplier range removed its lead on the fresh set. Second, for MPC-based flight control the fallback policy, solver acceptance rule, and reference information are part of the controller: without preview, a minority of episodes with fallback activity dominated the input variation, whereas with preview no fallback occurred at all on the figure-eight task. Third, a controller that completes a maneuver with instantaneous actuation can fail once measured computation delays are applied; P99 call time and the deadline policy should therefore be reported alongside accuracy, and real-time MPC designs should bound the iteration tail (e.g., by iteration caps tied to the deadline, delay-compensated prediction~\cite{findeisen2004}, or embedded solvers~\cite{verschueren2022,ferreau2014}). On a Raspberry Pi~5, a 100-iteration cap was enough to remove every deadline miss in the tested nominal step profiles.

\begin{table}[!t]\caption{Reporting checklist for quadrotor controller comparisons derived from this study.}\label{tab:checklist}\centering\footnotesize
	\begin{tabular}{p{3.6cm}p{4.2cm}}\toprule
		Report & Evidence in this study\\\midrule
		Tuning budget, search space, and selection data for every method & Ordering change, reproduced on fresh draws (Tables~\ref{tab:heldout}, \ref{tab:fresh})\\
		Reference information (preview, feedforward) per method & Preview: 0.169$\to$0.003 m (smooth transition); no fallback (figure-eight)\\
		Completion and failure counts with exact intervals before accuracy & 0/50 primary vs.\ 2/50 fresh (fixed BL-SMC)\\
		Solver acceptance rule and fallback frequency & 9.6\% fallback calls; 92.4\% of $V$\\
		Applied (not requested) input variation, including its distribution & LMPC-F: median 1.7, mean 194 s$^{-1}$\\
		Fallback policy and its interaction with authority & Raw 4/8 vs.\ projected 1/8 (step, single runs)\\
		Command-time distribution and deadline policy & Replay failures: 3/4 (desktop), 4/4 (Pi~5), 0/4 with a 100-iteration cap\\
		Actuator model in plant and predictor & Motor predictor shortens iteration tail\\
		\bottomrule
	\end{tabular}
\end{table}

\subsection{Scope and Limitations}\label{sec:scope}
\emph{Tuning and data.} The tuning results are conditional on a small budget (twelve candidates, two development episodes each, one validation episode per finalist), and the parameter spaces differ in expressiveness across controllers. The confirmation study shows that the mean-RMSE ordering is stable up to 48 candidates, but it does not establish optimal tuning; the single validation episode per finalist also makes selection noisy, as the geometric controller illustrates. The 50 primary evaluation realizations were inspected during earlier benchmark development; the 50 fresh realizations were not, and they reproduce the main conclusions. Neither protocol was externally preregistered. The BL-SMC vertical switching gain was searched over an absolute range of 1.5--4.5~m/s$^2$ instead of the 0.6--1.8 multiplier used for all other parameters. In a sensitivity check that restricts it to the multiplier range (sampled values 1.12--3.12~m/s$^2$) with the same candidates and selection rule, the same candidate is selected, now with a vertical gain of 2.89 instead of 4.01~m/s$^2$. Tuned BL-SMC then reaches 0.125~m on the primary realizations, still $-0.101$~m [$-0.109$, $-0.089$] below tuned PID, but 0.228~m on the fresh realizations, where its paired difference to tuned PID, $-0.003$~m [$-0.100$, $0.140$], no longer excludes zero. The first place of BL-SMC on the fresh set therefore depends on the wider vertical-gain range, whereas the primary-study result does not.

\emph{Architecture.} The shared 50~Hz rate, ideal sensing, hover-based prediction, and differences in reference construction and feedforward limit generalization to multi-rate flight stacks and aggressive flight~\cite{sun2022,faessler2017}. The primary evaluation uses MPC without preview to match the information available to the other controllers; the confirmation study shows that preview removes the fallback activity and reduces RMSE by 0.10~m, on the figure-eight task, so conclusions about MPC depend strongly on this choice. The tilt limits also differ: PID and the geometric controller clip the desired tilt at 0.55~rad, whereas LMPC-F imposes a 0.45~rad bound inside the QP. The controllers also differ in how they reject the 6-s force pulse. PID has integral action and BL-SMC a switching term, whereas LQR, LMPC-F, and the geometric controller have neither integral action nor a disturbance estimator, and the geometric baseline also uses zero desired angular velocity. The accuracy ordering therefore compares complete implementations, including their disturbance-rejection structure, and offset-free LQR or MPC variants were not tested. The motor variant changes state information, prediction model, and terminal cost simultaneously, and the one-step prediction errors are computed along different trajectories, so model fidelity alone is not isolated.

\emph{Fallback and constraints.} The projected fallback has no nonlinear recovery guarantee, and matched-state experiments would be required to isolate its causal effect. Soft-constraint feasibility does not imply satisfaction of the original tilt bound. The step-response mechanism cells are deterministic single runs, and their completion pattern has not been repeated over uncertainty or noise realizations. The share of LMPC-F infeasibility caused by carrying an unclipped fallback request into the next QP (Section~\ref{sec:results}) has not been quantified.

\emph{Timing.} Delays were measured on one desktop and one Raspberry Pi~5, both running general-purpose operating systems without real-time scheduling, and were replayed offline. The replay is exogenous: each trace is applied by call index, so the solve times are those of the delay-free trajectory and do not respond to the state changes that the delays cause. Only one deadline policy, holding the previous request, was tested; applying the LQR fallback or the best available iterate on a miss are alternatives. The replay also omits CPU queuing, sensor and communication latency, and the overhead of a full flight stack, and the iteration-cap results come from a single abrupt-step maneuver with three repeats. Because a few slow solves set P99 and these depend on platform-specific floating-point behavior, absolute timings do not transfer between machines, and none of the results are worst-case execution-time guarantees.

Future work will include flatness-based baselines with full angular-velocity feedforward~\cite{mellinger2011,faessler2018}, offset-free LQR and MPC variants, replication of the mechanism cells over uncertainty realizations, an endogenous timing model driven by the iteration counts of each replay, alternative deadline policies, multi-episode validation during selection, realistic sensing and wind, deadline-aware MPC evaluated over the full evaluation set and both fallback policies, compiled embedded solvers (e.g., OSQP code generation or acados~\cite{verschueren2022}) on a real-time kernel, and hardware-in-the-loop execution on a flight computer.

\section{Conclusion}\label{sec:conclusion}
Under a common nonlinear plant with rotor lag and saturation, tuning, reference information, fallback design, and command timing each materially changed the comparison of six quadrotor controller implementations. A matched twelve-candidate search improved every implementation and changed their mean-RMSE ordering, mainly by moving BL-SMC from last to first; the change recurred on 50 fresh realizations and did not change with a fourfold larger candidate-count budget, although BL-SMC's lead on the fresh set depends on the wider range searched for one of its gains. Reference preview reduced smooth-maneuver MPC error by more than an order of magnitude, whereas in single deterministic abrupt-step runs the projected fallback completed only in combination with preview, motor prediction, and reduced authority. The large input variation of LMPC-F was concentrated within fallback intervals in a minority of episodes and vanished when preview was enabled on the figure-eight task, and replaying measured command delays caused three of four nominal MPC profiles that had completed to terminate (all four on a Raspberry Pi~5), whereas a deadline-aware cap of 100 solver iterations prevented these failures in the tested nominal timing profiles. Controller comparisons should therefore report tracking accuracy together with tuning effort, failure counts, optimizer acceptance and fallback activity, applied-input variation, and command timing, as summarized in Table~\ref{tab:checklist}.

\section*{Declarations}
\emph{Data and code:} code, configurations, logs, and timing traces are available at \url{https://github.com/Quang61123/quadrotor_benchmark}. \emph{Generative AI} tools were used for English-language editing and for assistance in checking research code for implementation errors. All model outputs, source code, numerical results, analyses, references, and scientific conclusions were independently verified and approved by the authors. The authors take full responsibility for the content of the manuscript. \emph{Competing interests:} none.

\bibliographystyle{IEEEtran} 
\bibliography{references}

\end{document}